\documentclass{article}

\usepackage{spconf,amsmath,graphicx}
\usepackage{algorithm}
\usepackage{algpseudocode}
\usepackage{booktabs}
\usepackage{cite}
\usepackage{comment}
\usepackage{tabularray}
\usepackage{color}
\usepackage{multirow}
\usepackage{subcaption}
\usepackage{lipsum}
\usepackage{makecell} 
\usepackage{amssymb}
\usepackage{amsmath}
\usepackage{colortbl}
\usepackage[table]{xcolor}
\usepackage{array}
\usepackage{enumitem}
\usepackage{flushend}

\usepackage{pifont}
\usepackage[hidelinks]{hyperref}
\ninept

\title{BAD: Bidirectional Auto-regressive Diffusion for Text-to-Motion Generation}
\name{\normalsize{Seyed Rohollah Hosseyni}$^{1}$, Ali Ahmad Rahmani$^{2}$, Seyed Jamal Seyedmohammadi$^{3}$, Sanaz Seyedin$^{1}$, Arash Mohammadi$^{3}$}
\address{$^{1}$\normalsize{Dept. of Electrical Engineering, Amirkabir University of Technology (AUT), Tehran, Iran} \\
	$^{2}$\normalsize{Dept. of Electrical Engineering, Iran University of Science and Technology (IUST), Tehran, Iran} \\
	$^{3}$\normalsize{Concordia Institute of Information Systems Engineering (CIISE), Concordia University, Montreal, Canada}}

\begin{document}
%
\maketitle

\begin{abstract}
Autoregressive models excel in modeling sequential dependencies by enforcing causal constraints, yet they struggle to capture complex bidirectional patterns due to their unidirectional nature. In contrast, mask-based  models leverage bidirectional context, enabling richer dependency modeling. However, they often assume token independence during prediction, which undermines the modeling of  sequential dependencies. Additionally, the corruption of sequences through masking or absorption can introduce unnatural distortions, complicating the learning process. To address these issues, we propose Bidirectional Autoregressive Diffusion (BAD), a novel approach that unifies the strengths of autoregressive and mask-based generative models. BAD utilizes a permutation-based corruption technique that preserves the natural sequence structure while enforcing causal dependencies through randomized ordering, enabling the effective capture of both sequential and bidirectional relationships. Comprehensive experiments show that BAD outperforms autoregressive and mask-based models in text-to-motion generation, suggesting a novel pre-training strategy  for sequence modeling.  The codebase for BAD is available on~\url{https://github.com/RohollahHS/BAD}.
\end{abstract}
\begin{keywords}
Motion Generation - Autoregressive Models - Mask-based Generative Models - Diffusion Models
\end{keywords}

\vspace*{-.1in}
\section{Introduction} \label{sec:intro}
\vspace*{-.1in}
Text-to-motion generation~\cite{Ren:2023, petrovich2022temos, chen2023executing, guo2022generating, zhang2022motiondiffuse} is an emerging field that integrates natural language processing with 3D human motion synthesis, offering substantial potential for applications in gaming, film industry, virtual reality, and robotics~\cite{azadi2023make, 10446237, tevet2022motionclip}. This task is inherently challenging due to the difficulty of mapping discrete textual descriptions into continuous, high-dimensional motion data. To address this challenge, Vector-Quantized Variational Autoencoders (VQ-VAEs)~\cite{oord2018neural} have proven to be particularly effective  in text-to-motion generation~\cite{guo2022tm2t, zhang2023generating, pinyoanuntapong2024mmm}. Typically, a two-stage approach is followed where a VQ-VAE is first trained to transform continuous motion data into discrete motion tokens. In the second stage and to model the distribution of motion tokens in discrete space, either autoregressive or denoising models are employed. Nevertheless, despite their effectiveness, each category has its inherent limitations as outlined below.

\vspace{.05in}
\noindent
\textbf{\textit{Literature Review:}} Autoregressive models excel at capturing and leveraging sequential dependencies on various modalities \cite{van2016pixel, tian2024visual, hoogeboom2022autoregressive, van2016wavenet, kalchbrenner2018efficient, brown2020language} due to their reliance on the causality of the input. In these models, each token is predicted based on previously generated tokens, allowing the model to naturally learn the progression and relationship between consecutive tokens. Employing autoregressive models to learn discrete motion sequences has led to significant improvements in text-to-motion generation, generating high-fidelity and coherent motion sequences~\cite{zhang2023generating, zhong2023attt2m, jiang2024motiongpt}. The unidirectional nature of these models, however, limit their ability to fully capture deep bidirectional context, as they only consider the preceding tokens lacking insight into the future ones.

Conversely, denoising models, particularly mask-based generative models~\cite{devlin2019bertpretrainingdeepbidirectional} or absorbing diffusion models~\cite{austin2021structured}, leverage both preceding and subsequent contexts to capture rich bidirectional relationships, eliminating unidirectional bias. By adopting this approach, mask-based motion models~\cite{pinyoanuntapong2024mmm, guo2024momask} enhance the generation of complex motion sequences over autoregressive motion models.
Mask-based generative models, however, assume that masked tokens are conditionally independent~\cite{yang2019xlnet}, meaning predictions do not account for potential dependencies between masked tokens, which can result in suboptimal predictions. Furthermore, the corruption process in these models involves transitioning certain tokens in the input sequence to a \texttt{[MASK]} token or an absorbed state. Encoding a portion or the entire sequence into a fully masked (absorbed) form is an unnatural process, which distorts the sequence and complicates the task of learning the corresponding reverse noise-to-data mapping.

\vspace{.05in}
\noindent
\textbf{\textit{Contributions:}} Motivated by the aforementioned limitations of autoregressive and mask-based generative models, we propose the Bidirectional Autoregressive Diffusion (BAD) framework, a novel pretraining strategy for sequence modeling that unifies the strengths of both autoregressive and mask-based generative models. We evaluate BAD in the context of text-to-motion generation in a two-stage process. In the first stage, we train a motion tokenizer based on the conventional VQ-VAE to convert motion sequences into discrete representations using a learned codebook. In the second stage, the proposed BAD is used to train a transformer architecture. This process begins with a novel corruption method designed based on permutation operation. Specifically, we utilize multiple different mask tokens (absorbed states) and a random ordering to systematically corrupt the sequence, resulting in a more natural corrupted sequence. After randomly masking a portion of the motion sequence, a hybrid attention mask, which integrates a permuted causal attention mask and a bidirectional attention mask, is constructed to determine the dependencies among input tokens. The permuted causal attention mask enforces each masked token to learn its causal dependencies on others, while the bidirectional attention mask ensures that all tokens can attend to both preceding and subsequent unmasked tokens, therefore, enriching the model's capacity to capture sequential dependencies and deep bidirectional context. 

Although our primary goal is to address issues related to autoregressive and mask-based generative models using the proposed BAD framework, we demonstrate that by using a simple VQ-VAE as our motion tokenizer in the first stage, our model can achieve competitive or superior results compared to models employing advanced VQ-VAEs, such as Residual Vector Quantization (RVQ)~\cite{zeghidour2021soundstream}, used in \cite{guo2024momask, pinyoanuntapong2024bamm}. In RVQ-VAE, multiple layers of vector quantization are applied sequentially, with each layer encoding residual information not captured by the preceding layers. Such a hierarchical approach significantly enhances the performance of the motion tokenizer and, consequently, that of the overall framework. Using RVQ-VAE, however, often requires training multiple transformers and incurs additional network calls during inference to predict motion tokens associated with the residual layers in the second stage. These will greatly increase the computational complexity and training time of the underlying model. In contrast, the proposed framework, which uses a simple VQ-VAE as its motion tokenizer, requires training only a single transformer in the second stage. Furthermore, it requires far fewer number of network calls during inference, while achieving comparable results to RVQ-VAE-based models. In summary, the paper makes the following key contributions:
\begin{itemize}
\item Introduction of BAD framework, which integrates the bidirectional capabilities of mask-based generative models with the causal dependencies inherent in autoregressive modeling.
\item Introduction of a novel corruption (diffusion) technique for discrete data in the context of text-to-motion generation. The proposed technique, unlike prior works, preserves the sequential nature of data, facilitating a more natural learning process.
\end{itemize}
Extensive experiments are performed based on widely recognized text-to-motion  HumanML3D~\cite{guo2022generating} and KIT-ML~\cite{plappert2016kit} datasets. Our results demonstrate the superiority of the proposed BAD framework against autoregressive and mask-based motion baseline models. Specifically, we improve the Frechet Inception Distance (FID) of~\cite{pinyoanuntapong2024mmm}, a mask-based generative motion model, from $0.089$ to $0.049$ on HumanML3D and from $0.316$ to $0.221$ on  KIT-ML dataset, while maintaining a similar model size and design choices. We also show that BAD achieves comparable results to methods utilizing advanced motion tokenizers, highlighting its efficiency and effectiveness. Finally, we show that BAD performs quite well on other tasks, such as text-guided motion inpainting and outpainting.

\vspace*{-.1in}
\section{The BAD Framework} \label{sec:method}

Our objective is to develop a text-to-motion generation framework that, given a textual description, generates coherent and complex human motion sequences. As illustrated in Fig.~\ref{fig:framework}, the proposed framework consists of two main components: (i) A motion tokenizer (Section~\ref{sec:tokenizer}), and; (ii) A conditional transformer (Section~\ref{sec:transformer}). The motion tokenizer converts raw 3D motion into discrete tokens, while the transformer predicts the original tokens from a corrupted sequence, conditioned on a text prompt. During inference (Section~\ref{sec:method-inference}), given a text prompt, the transformer starts with a noise vector $ \mathbf{z} $ and iteratively denoises it to generate a motion sequence.

\vspace*{-.15in}
\subsection{Motion Tokenizer} \label{sec:tokenizer}
\vspace*{-.075in}
The motion tokenizer, illustrated in Fig.~\ref{fig:framework}(a), comprises of an encoder and a decoder. Consider a raw motion sequence \( {F} = \{ {f}_1, {f}_2, \dots, {f}_\tau \} \) with $\tau$ frames, where \( {f}_t \in \mathbb{R}^D \) denotes the motion vector with dimensionality of $D$ at frame \( t \).  The encoder maps the raw motion sequence \( {F} \) into a continuous latent space, yielding \( {E} = \{ {e}_1, {e}_2, \dots, {e}_{T} \} \) with  $ T = \tau / l $, where \( {e}_t \in \mathbb{R}^d \) is the latent vector with dimensionality of $d$, and \( l \) is the temporal downsampling rate. To obtain a discrete representation, each latent vector \( {e}_t \) is mapped to the nearest vector in a learned codebook \( \mathcal{C} = \{ {c}_k \in \mathbb{R}^d \mid k = 1, 2, \dots, K \} \), where \( K \) is the number of codebook entries. The quantized latent vector is defined as \( {x}_t = \text{Quantize}({e}_t) = {c}_k \), where \( k = \arg\min_j \| {e}_t - {c}_j \| \).  Finally, the  decoder receives the quantized or discrete motion sequence \( {X} = \{ {x}_1, {x}_2, \dots, {x}_{T} \} \) to reconstruct the raw motion sequence  \( {\hat{F}} = \{ {\hat{f}}_1, {\hat{f}}_2, \dots, {\hat{f}}_\tau \} \). The objective function for training the VQ-VAE is given by
\begin{equation}
	L_{vq} = \Vert {F} - \hat{{F}} \Vert_1 + \Vert \text{sg}[{E}] - {X} \Vert_2 + \beta \Vert {E} - \text{sg}[{X}] \Vert_2,
\end{equation}
where \( \beta \) controls the commitment loss, and \( \text{sg}(.) \) denotes the stop-gradient operation.

\vspace*{-.15in}
\subsection{Conditional Mask-Based Transformer}\label{sec:transformer}
\vspace*{-.075in}
Our transformer is designed to model the distribution of discrete motion tokens conditioned on a given textual description. The associated textual description is first processed through a pre-trained Contrastive Language-Image Pretraining (CLIP) model~\cite{radford2021learning}, yielding sentence and word embeddings that capture both global and local relationships between the text and motion sequence. Sentence embedding is prepended to the motion sequence, and word embeddings are integrated via cross-attention at the begging of the transformer.

\vspace{.05in}
\noindent
\textbf{\textit{Corruption Process:}} Let $ \mathcal{Z}_{T} $ denote the set of all possible permutations of the sequence $[1, 2, \dots, T]$, where $T$ is the sequence length. We define the $p$-th element of a permutation $ \mathbf{z} \in \mathcal{Z}_{T} $ as $ z_{p} $, with the first $p$ elements as $ \mathbf{z}_{\leq p} $ and the last $T\!\!-\!p\!+\!\!1$ elements as $ \mathbf{z}_{\geq p} $.

\begin{figure}[t]
\centering
\includegraphics[width=0.4\textwidth]{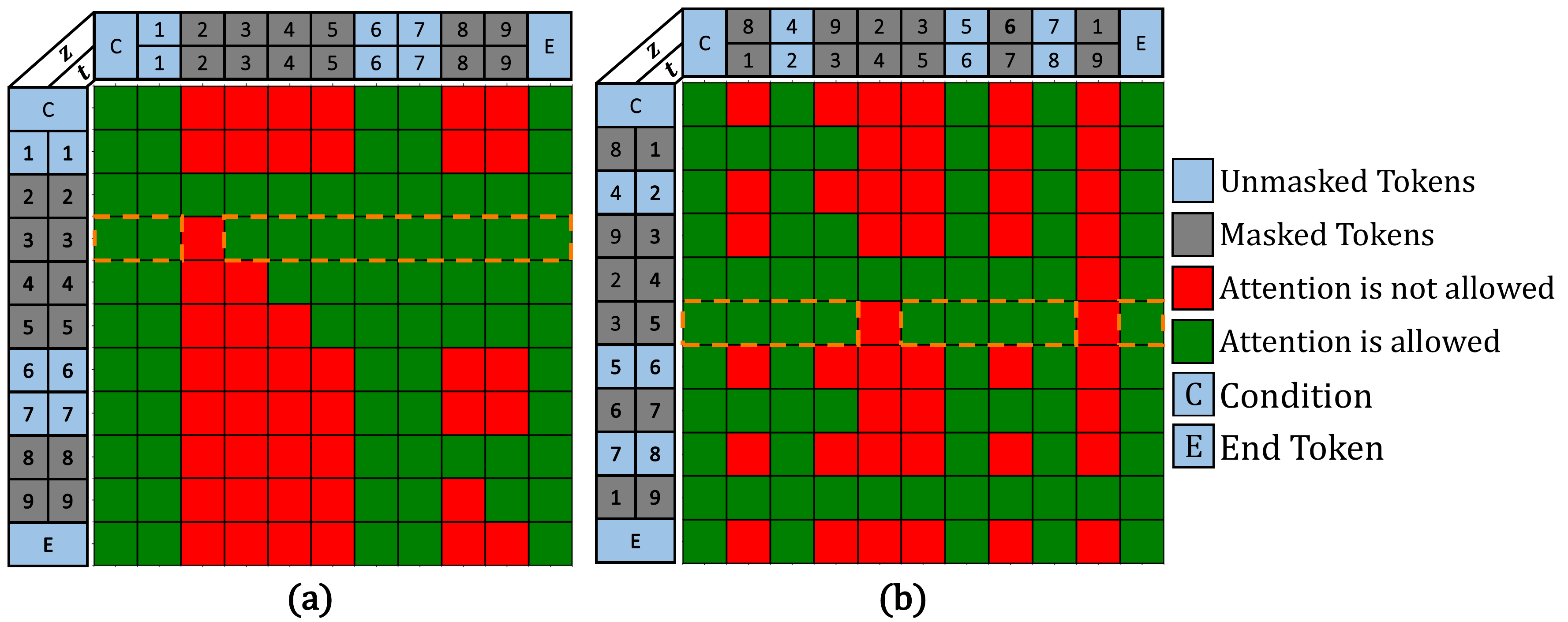}
\vspace*{-0.1in}\caption{\footnotesize Examples of two different hybrid attention masks. \( \mathbf{z} \) represents a random ordering $\mathbf{z} \sim \mathcal{Z}_{T}$, while \( t \) denotes time. Each mask token attends to the last $T\!\!-\!p\!+\!\!1$ mask tokens $\mathbf{m}_{\mathbf{z} \geq p}$ and unmasked tokens. For example, \textcolor{orange}{orange} cells indicate tokens that the third mask token, $m_{z_{3}}$, can attend to, including unmasked tokens and the existing $\mathbf{m}_{\mathbf{z} \geq 3}$ mask tokens.}
	\label{fig:multi-mask}
\end{figure}

Given a discrete motion sequence \( X = (x_1, x_2, \dots, x_T) \), we first randomly select \( n_m \) candidate motion tokens to be masked, resulting in a corrupted motion sequence composed of masked tokens \( X_m \) and unmasked tokens \( X_u \). Using \( X_u \), a bidirectional attention mask \( \text{att}_{\text{bi}} \) is created, which allows all tokens to attend to unmasked tokens from both directions. Next, we sample a random ordering \( \mathbf{z} \sim \mathcal{Z}_{T} \) to determine the order of all \( T \) mask tokens \( \mathbf{m} = (m_1, m_2, \dots, m_T) \), where \( m_i \in \mathbb{R}^d \), from our Maskbook. Using $\mathbf{z}$,  the corresponding permuted causal attention mask \( \text{att}_{\text{per}} \) is created, which enforces that each mask token \( m_{z_{p}} \) at position \( z_{p} \) can only attend to the last $T\!\!-\!p\!+\!\!1$ mask tokens, denoted by \( \mathbf{m}_{ \mathbf{z} \geq p } \). Finally, the candidate masked tokens \( X_m \) are replaced with \( n_m \) randomly selected mask tokens, and the hybrid attention mask is constructed as \( \text{att}_{\text{hyb}} = \text{att}_{\text{bi}} + \text{att}_{\text{per}} \). The hybrid attention mask ensures that mask tokens attend only to \( \mathbf{m}_{\mathbf{z} \geq p} \), maintaining causal dependencies similar to autoregressive models. Additionally, mask tokens can attend to unmasked tokens, while unmasked tokens only attend to each other. By attending to both the left and right unmasked tokens, our transformer effectively captures bidirectional context, similar to BERT~\cite{devlin2019bertpretrainingdeepbidirectional}. Fig.~\ref{fig:multi-mask} illustrates examples of the hybrid attention masks.

\begin{figure}[t!]
\centering
\includegraphics[width=0.45\textwidth]{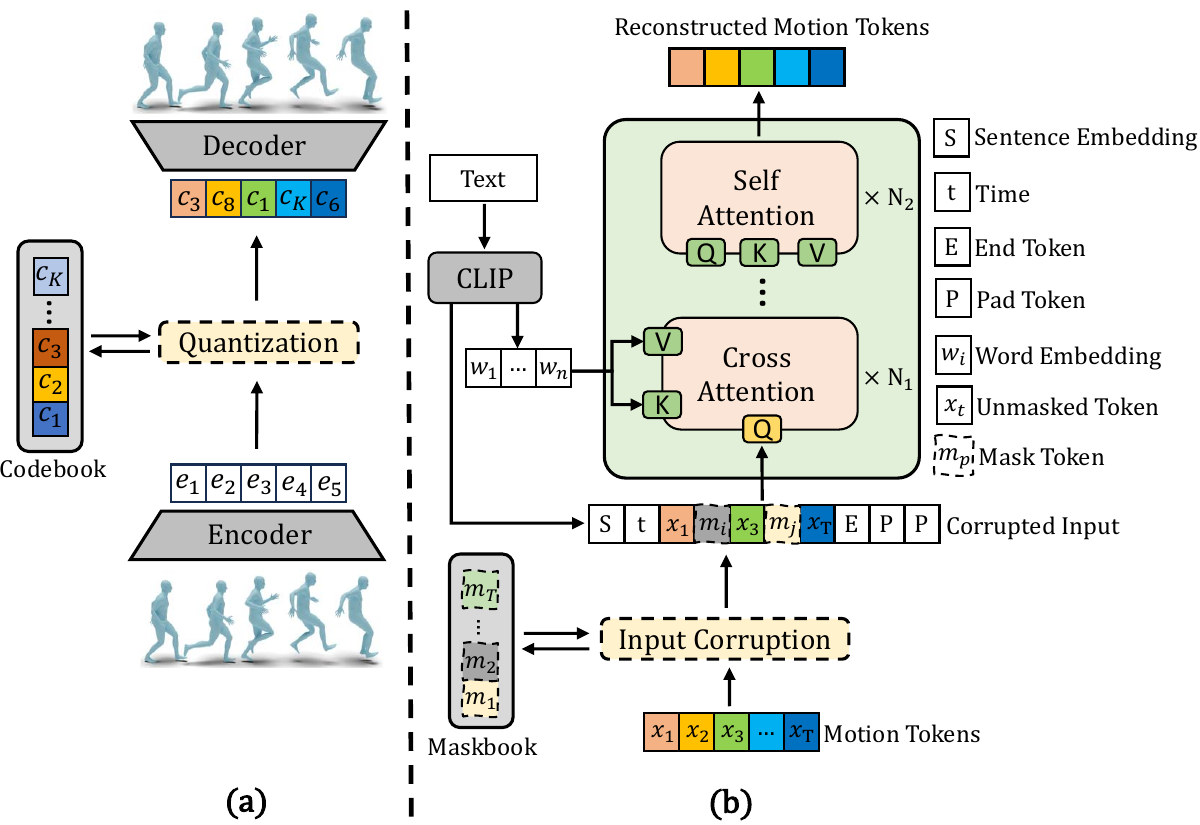}
\vspace*{-0.1in}\caption{\footnotesize Overall framework of our text-to-motion model. (a) Motion tokenizer, transforms  a raw 3D motion sequence into a sequence of discrete motion tokens. (b) The conditional mask-based transformer reconstructs original discrete motion tokens from a corrupted sequence conditioned on a text prompt.}
\label{fig:framework}
\end{figure}

\vspace{.05in}
\noindent
\textbf{\textit{Note:}} Following previous works, we use random replacement augmentation by replacing $c_r \times 100\%$ of  ground-truth motion tokens with random ones before masking, where $c_r \sim \textit{U}(0, 0.4)$. The number of tokens for masking, $n_{m}$, is also obtained as $c_m \times 100\%$ of the sequence length, where $c_m$ is sampled from $\textit{U}(0, 0.5)$ with a probability of $0.1$ or $\textit{U}(0.5, 1)$ with a probability of $0.9$. $n_m$ can also be prepended to the motion sequence, denoted as $time$ in Fig.~\ref{fig:framework}(b).

\vspace{.05in}
\noindent
\textbf{\textit{Objective Function:}} Our objective function is expressed as follows
\begin{equation}
	\underset{\theta}{\text{max}} \;\;\;\;\; \mathbb{E}_{ \mathbf{z}  \sim  \mathcal{Z}_{T}} \sum_{z_p=1}^{T} \; \left\{ \begin{array}{lcl}
		m^{\prime} \; \text{log} \; p_{\theta} ( \; x_{z_p} \; | \; \mathbf{m}_{\mathbf{z} \geq p } \; , X_{u}) \\
		(1-m^{\prime}) \; \text{log} \; p_{\theta} ( \; x_{z_p} \; | \; X_{u})
	\end{array}\right.\label{eq:ObjFun}
\end{equation}
where $m^{\prime}=1$ if $x_{z_p}$ is masked. The first part of Eq.~\eqref{eq:ObjFun} aligns with the autoregressive objective, thus avoiding the independence assumption of masked tokens during prediction.  For the sake of simplicity, other conditions, including \(S\) (sentence embedding), \(W\) (word embeddings), and \(t\) (time), have been omitted from Eq.~\eqref{eq:ObjFun}.

\setlength{\textfloatsep}{0pt}
\vspace*{-.15in}
\subsection{Inference} \label{sec:method-inference}
\vspace*{-.075in}

Given the permutation-based nature of our procedure, the proposed model can be trained under either \( \mathbf{m}_{ \mathbf{z} \geq p} \) or \( \mathbf{m}_{ \mathbf{z} \leq p} \) condition. Under \( \mathbf{m}_{ \mathbf{z} \leq p} \) condition, the mask tokens should attend to the first $p$ mask tokens. Different generation methods can then be applied using the same trained model. In this paper, we demonstrate two of such methods. Each generation method can employ parallel decoding, where the transformer decodes all mask tokens while selectively masking others based on a cosine scheduling function, \( n_m = T \cos\left(\frac{1}{2} \pi i / I\right) \), where \(i\) and \(I\) represent the current iteration and the total number of iterations, respectively. Initially, a high masking ratio is applied, masking most of the motion tokens. As the generation process progresses, the masking ratio is gradually reduced, increasing the available context. This increasing context allows the model to infer the remaining masked tokens more accurately. To determine the number of mask tokens \( n_m \) at each iteration \( i \), we need the sequence length \( T \). This sequence length can also be masked and learned by the model, which requires minor modifications. However, since our goal is to propose BAD, we aim to keep everything simple. Alternatively, one can use a length estimator or pre-specify \( T \).

\vspace*{-.12in}
\subsubsection{Order-Agnostic Autoregressive Sampling (OAAS)} \label{sec:order-agnostic-ar}
\vspace*{-.075in}
In this approach, we first sample a random ordering \( \mathbf{z} \sim \mathcal{Z}_{T} \) to create mask tokens and the corresponding permuted causal attention mask. Decoding begins from \( m_{z_{1}} \), allowing this token to attend to all other mask tokens \( \mathbf{m}_{\mathbf{z} \geq 1} \) and use the rich information they captured during training. In the subsequent iterations, the hybrid attention mask is updated, and mask tokens are allowed to attend only to the last $T\!\!-\!p\!+\!\!1$ mask tokens \( \mathbf{m}_{\mathbf{z} \geq p} \) and unmasked tokens.  This iterative process continues until all tokens are decoded. Alternatively, for the model under \( \mathbf{m}_{\mathbf{z} \leq p} \) condition, decoding starts from \( m_{z_{T}} \).

\vspace*{-.12in}
\subsubsection{Confidence-Based Sampling (CBS)} \label{sec:confidence}
\vspace*{-.075in}
This approach also initiates generation from randomly ordered mask tokens based on a random ordering \( \mathbf{z} \sim \mathcal{Z}_{T} \).  During decoding, tokens predicted with high confidence are retained, while lower-confidence tokens are masked for further processing. This ensures that the sequence benefits from the most reliable predictions, potentially enhancing the quality of the generated sequence.

\vspace*{-.1in}
\section{Experiments}\label{sec:experiments}
\vspace*{-.1in}

\textbf{\textit{Datasets:}} We conducted experiments using two widely recognized text-to-motion datasets: (i) HumanML3D dataset \cite{guo2022generating}, a large-scale dataset with 14,616 motion sequences and $44,970$ textual descriptions, combining data from AMASS~\cite{mahmood2019amass}, and HumanAct12~\cite{guo2020action2motion}, and; (ii) KIT-ML dataset~\cite{plappert2016kit}, a smaller benchmark with $3,911$ motion sequences and $6,278$ textual annotations, sourced from the KIT~\cite{mandery2015kit} and CMU~\cite{cmu_mocap} motion databases. 

\vspace{.05in}
\noindent
\textbf{\textit{Evaluation Metrics:}} For evaluations, we use standard metrics from previous works~\cite{guo2022generating}, leveraging pre-trained models to encode text and motion features. We assess the alignment between generated motions and text prompts using R-Precision, reporting Top-1, Top-2, and Top-3 accuracies. To evaluate motion quality, we calculate Frechet Inception Distance (FID) to measure the distributional difference between generated and real motion features. We also assess diversity by computing the average Euclidean distance between randomly selected pairs of generated motions and Multimodality by measuring variance across multiple motions generated from the same prompt. These metrics together provide a comprehensive understanding of the quality and diversity of the generated motions relative to the text prompt. 

\vspace{.05in}
\noindent
\textbf{\textit{Implementation Details:}} Following~\cite{pinyoanuntapong2024mmm}, we use a simple VQ-VAE motion tokenizer with a codebook size of $8,192$ and a dimension of $32$, along with a temporal downsampling rate of $l = 4$. For training, motion sequences from  HumanML3D and KIT-ML datasets are truncated to a length of $\tau = 64$. The model is optimized using  AdamW optimizer with  $[\beta_1, \beta_2] = [0.9, 0.99]$, a batch size of $256$, and an exponential moving average constant $\lambda = 0.99$. We initially train for $200$K iterations using a learning rate of $2 \times 10^{-4}$, then continue for another $100$K iterations with a reduced learning rate of $1 \times 10^{-5}$. In the second stage, we use a transformer~\cite{vaswani2017attention} consisting of $18$ layers, each with a dimension of $1,024$ and $16$ attention heads. The first two layers are cross-attention layers, while the rest are self-attention layers. The transformer is also trained using AdamW with $[\beta_1, \beta_2] = [0.5, 0.99]$ and a batch size of $128$. The learning rate is initially set at $2 \times 10^{-4}$ for the first $150$K iterations and is subsequently decayed to $1 \times 10^{-5}$ for the rest of the training. 

\vspace*{-.15in}
\subsection{Comparison with state-of-the-art approaches}
\vspace*{-2mm}
\begin{table*}[t]
	\centering
	\resizebox{0.73\textwidth}{!}{
		\begin{tabular}{clccccccc}
			\Xhline{1.5pt}
			\rule{0pt}{2.5ex} \multirow{2}{*}{Dataset} & \multirow{2}{*}{Methods} & \multicolumn{3}{c}{R-Precision $\uparrow$} & \multirow{2}{*}{FID $\downarrow$} & \multirow{2}{*}{MM-Dist $\downarrow$} & \multirow{2}{*}{Diversity $\uparrow$} & \multirow{2}{*}{MModality $\uparrow$} \\ \cline{3-5} 
			\rule{0pt}{2.5ex} & & Top-1 & Top-2 & Top-3 & & & &  \\ \hline  
			\rule{0pt}{2.5ex} \multirow{10}{*}{\rotatebox{90}{HumanML3D}} &  Real & $0.511^{\pm .003}$ & $0.703^{\pm .003}$ & $0.797^{\pm .002}$ & $0.002^{\pm .000}$ & $2.974^{\pm .008}$ & $9.503^{\pm .065}$ & - \\ 
			& VQ-VAE & $0.505^{\pm .002}$ & $0.697^{\pm .003}$ & $0.790^{\pm .002}$ & $0.085^{\pm .001}$ & $3.031^{\pm .009}$ & $9.650^{\pm .073}$ & - \\ \cline{2-9}
			\rule{0pt}{2.5ex}
			& MDM \cite{tevet2022humanmotiondiffusionmodel} & $0.320^{\pm .005}$ & $0.498^{\pm .004}$ & $0.611^{\pm .007}$ & $0.544^{\pm .044}$ & $5.566^{\pm .027}$ & $9.559^{\pm .086}$ & $2.799^{\pm .072}$ \\
			& MotionGPT \cite{jiang2024motiongpt} &  $0.435^{\pm .003}$ & $0.607^{\pm .002}$ & $0.700^{\pm .002}$ & $0.160^{\pm .008}$ & $3.700^{\pm .009}$ & $9.411^{\pm .081}$ & $\textcolor{black}{\mathbf{3.437}^{\pm \mathbf{.091}}}$ \\
			& T2M-GPT \cite{zhang2023generating} & $0.491^{\pm .003}$ & $0.680^{\pm .003}$ & $0.775^{\pm .002}$ & $0.116^{\pm .004}$ & $3.118^{\pm .011}$ & $\textcolor{black}{\mathbf{9.761}^{\pm \mathbf{.081}}}$ & $1.856^{\pm .011}$ \\
			& AttT2M \cite{zhong2023attt2m} & $0.499^{\pm .003}$ & $0.690^{\pm .002}$ & $0.786^{\pm .002}$ & $0.112^{\pm .006}$ & $3.038^{\pm .007}$ & $\textcolor{black}{{\underline{9.700}}^{\pm {.090}}}$ & $\textcolor{black}{{\underline{2.452}}^{\pm {.051}}}$ \\
			& MMM \cite{pinyoanuntapong2024mmm} & ${\underline{0.515}}^{\pm .002}$ & ${\underline{0.708}}^{\pm .002}$ & ${\underline{0.804}}^{\pm .002}$ & $0.089^{\pm .005}$ & $\underline{2.926}^{\pm .007}$ & $9.577^{\pm .050}$ & $1.226^{\pm .035}$ \\   \cline{2-9}
			\rule{0pt}{2.5ex} & BAD (CBS \ref{sec:confidence}) & $0.511^{\pm .002}$ & $0.704^{\pm .002}$ & $0.800^{\pm .002}$ & ${\mathbf{0.049}}^{\pm \mathbf{.003}}$ & $2.957^{\pm .006}$ & $9.688^{\pm .089}$ & $1.119^{\pm .042}$ \\
			& BAD (OAAS \ref{sec:order-agnostic-ar}) & $\textcolor{black}{{\mathbf{0.517}}^{\pm \mathbf{.002}}}$ & $\textcolor{black}{\mathbf{0.713}^{\pm \mathbf{.003}}}$ & $\textcolor{black}{\mathbf{0.808}^{\pm \mathbf{.003}}}$ & $\textcolor{black}{{\underline{0.065}}^{\pm {.003}}}$ & $\textcolor{black}{\mathbf{2.901}^{\pm \mathbf{.008}}}$ & $9.694^{\pm .068}$ & $1.194^{\pm .044}$ \\  \Xhline{1.5pt}
			\rule{0pt}{2.5ex} \multirow{11}{*}{{\rotatebox{90}{KIT-ML}}} & Real & $0.424^{\pm .005}$ & $0.649^{\pm .006}$ & $0.779^{\pm .006}$ & $0.031^{\pm .004}$ & $2.788^{\pm .012}$ & $11.080^{\pm .097}$ & - \\ 
			& VQ-VAE & $0.400^{\pm .006}$ & $0.619^{\pm .006}$ & $0.746^{\pm .007}$ & $0.437^{\pm .010}$ & $2.981^{\pm .017}$ & $11.093^{\pm .095}$ & - \\ \cline{2-9}
			\rule{0pt}{2.5ex}
			& MDM \cite{tevet2022humanmotiondiffusionmodel} & $0.164^{\pm .004}$ & $0.291^{\pm .004}$ & $0.396^{\pm .004}$ & $0.497^{\pm .021}$ & $9.191^{\pm .022}$ & $10.85^{\pm .109}$ & $1.907^{\pm .214}$ \\
			& MotionGPT \cite{jiang2024motiongpt} &  $0.366^{\pm .005}$ & $0.558^{\pm .004}$ & $0.558^{\pm .005}$ & $0.510^{\pm .016}$ & $3.527^{\pm .021}$ & $10.35^{\pm .084}$ & $\textcolor{black}{\mathbf{2.328}^{\pm .117}}$ \\
			& T2M-GPT \cite{zhang2023generating} & $0.402^{\pm .006}$ & $0.619^{\pm .005}$ & $0.737^{\pm .006}$ & $0.717^{\pm .041}$ & $3.053^{\pm .026}$ & $10.86^{\pm .094}$ & $1.912^{\pm .036}$ \\
			& AttT2M \cite{zhong2023attt2m} & $\underline{0.413}^{\pm .006}$ & ${\mathbf{0.632}}^{\pm \mathbf{.006}}$ & $\mathbf{0.751}^{\pm \mathbf{.006}}$ & $0.870^{\pm .039}$ & $3.039^{\pm .021}$ & $\underline{10.96}^{\pm .123}$ & $\textcolor{black}{{\mathbf{2.281}}^{\pm \mathbf{.047}}}$ \\
			& MMM \cite{pinyoanuntapong2024mmm} & $0.404^{\pm .005}$ & $0.621^{\pm .005}$ & $0.744^{\pm .004}$ & $0.316^{\pm .028}$ & ${\underline{2.977}}^{\pm .019}$ & $10.91^{\pm .101}$ & $1.232^{\pm .039}$ \\   \cline{2-9}
			\rule{0pt}{2.5ex} & BAD (CBS \ref{sec:confidence}) & $0.408^{\pm .004}$ & $0.612^{\pm .007}$ & $0.734^{\pm .007}$ & ${\underline{0.246}}^{\pm .019}$ & $3.100^{\pm .021}$ & $10.874^{\pm .083}$ & $1.485^{\pm .059}$ \\
			& BAD (OAAS \ref{sec:order-agnostic-ar}) & ${\mathbf{0.417}}^{\pm \mathbf{.006}}$ & ${{\underline{0.631}}}^{\pm {.006}}$ & $\textcolor{black}{{{{\underline{0.750}}}^{\pm {.006}}}}$ & ${\mathbf{0.221}}^{\pm \mathbf{.012}}$ & ${\mathbf{2.941}}^{\pm \mathbf{.025}}$ & $\textcolor{black}{{\mathbf{11.000}}^{\pm \mathbf{.100}}}$ & $1.170^{\pm .047}$ \\ 
\Xhline{1.5pt}
\end{tabular}
}
\vspace*{-0.1in} \caption{\footnotesize Quantitative evaluation on  HumanML3D and KIT-ML test sets. Best results are in \textbf{bold}, with second-best \underline{underlined}. The evaluation is repeated 20 times for each metric, and the mean is reported along with the 95\% confidence interval, denoted by $\pm$.} 
\label{tab:no-rvq}
\vspace*{-0.1in}
\end{table*}
\begin{table}[t]
	\centering
	\resizebox{0.5\textwidth}{!}{
		\begin{tabular}{clccccccc}
			\Xhline{1.5pt}
			\rule{0pt}{2.5ex} \multirow{2}{*}{Dataset} & \multirow{2}{*}{Methods} & \multicolumn{3}{c}{R-Precision $\uparrow$} & \multirow{2}{*}{FID $\downarrow$} & \multirow{2}{*}{MM-Dist $\downarrow$} & \multirow{2}{*}{Diversity $\uparrow$} & \multirow{2}{*}{MModality $\uparrow$} \\ \cline{3-5} 
			\rule{0pt}{2.5ex} & & Top-1 & Top-2 & Top-3 & & & &  \\ \hline  
			\rule{0pt}{2.5ex} \multirow{4}{*}{{\rotatebox{90}{{\footnotesize HumanML3D}}}} & MoMask \cite{guo2024momask} & ${{0.521}}$ & ${{0.713}}$ & $0.807$ & ${{{0.045}}}$ & $2.958$ & - & $1.241$ \\
			& BAMM \cite{pinyoanuntapong2024bamm} & ${{0.525}}$ & ${{0.720}}$ & ${{0.814}}$ & $0.055$ & ${{2.919}}$ & ${{9.717}}$ & $1.687$ \\ \cline{2-9}
			\rule{0pt}{3ex} & BAD (CBS) & $0.511$ & $0.704$ & $0.800$ & $0.049$ & $2.957$ & $9.688$ & $1.119$ \\
			& BAD (OAAS) & $0.517$ & $0.713$ & $0.808$ & $0.065$ & $2.901$ & $9.694$ & $1.194$ \\  \Xhline{1.5pt}
			\rule{0pt}{2.5ex} \multirow{4}{*}{{\rotatebox{90}{\footnotesize KIT-ML}}} & MoMask \cite{guo2024momask} & $0.433$ & $0.656$ & $0.781$ & $0.204$ & $2.779$ & - & $1.131$  \\ 
			& BAMM \cite{pinyoanuntapong2024bamm} & $0.438$ & $0.661$ & $0.788$ & $0.183$ & $2.723$ & $11.008$ & $1.609$ \\ \cline{2-9}
			\rule{0pt}{2.5ex} & BAD (CBS) & $0.408$ & $0.612$ & $0.734$ & $0.246$ & $3.100$ & $10.874$ & $1.485$ \\
			& BAD (OAAS) & $0.417$ & $0.631$ & $0.750$ & $0.221$ & $2.941$ & $11.000$ & $1.170$ \\ 
\Xhline{1.5pt}
\end{tabular}
}
\vspace*{-0.1in} \caption{\footnotesize Quantitative evaluation on  HumanML3D and KIT-ML test sets in comparison to RVQ-VAE-based models.} 
\label{tab:rvq}
\vspace*{-0.1in}
\end{table}
\begin{table}[t!]
	\centering
	\resizebox{0.9\columnwidth}{!}{
		\begin{tabular}{llcccc}
			\Xhline{1.5pt}
			\multirow{2}{*}{ Task}                                                                                                                                                                       & \multirow{2}{*}{ Method} & \multirow{2}{*}{{\begin{tabular}[c]{@{}l@{}}  {R-Precision}\\  { \hspace{0.2cm} Top-3 $\uparrow$}\end{tabular}}} & \multirow{2}{*}{ FID $\downarrow$} & \multirow{2}{*}{ MM-Dist $\downarrow$} & \multirow{2}{*}{ Diversity $\uparrow$} \\
			&                                               &                        &                                                         &                                                             &                                                             \\ \hline                                         
			\rule{0pt}{2.5ex} \multirow{3}{*}{{\begin{tabular}[c]{@{}l@{}}  {Temporal Inpainting}\\  {(In-betweening)}\end{tabular}}} &  Momask                  & $0.820$                & $\mathbf{0.040}$                                                 & $2.878$                                                     & $\mathbf{9.640}$                                                     \\
			&  BAMM                    & $\mathbf{0.821}$                & $0.056$                                                 & $\mathbf{2.863}$                                                     & $9.629$                                                     \\
			&  BAD                     & $0.810$                & $0.045$                                                 & $2.899$                                                     & $9.546$                                                     \\ \hline
			\rule{0pt}{2.5ex} \multirow{3}{*}{{\begin{tabular}[c]{@{}l@{}}  {Temporal}\\  {Outpainting}\end{tabular}}}  &  Momask                  & $0.818$                & $0.057$                                                 & $2.889$                                                     & $9.619$                                                     \\
			&  BAMM                    & $\mathbf{0.822}$                & $0.056$                                                 & $\mathbf{2.856}$                                                     & $\mathbf{9.659}$                                                     \\
			&  BAD                     & $0.800$                & $\mathbf{0.034}$                                                 & $2.961$                                                     & $9.579$                                                     \\ \hline
			\rule{0pt}{2.5ex} \multirow{3}{*}{Prefix}  &  Momask                  & $\mathbf{0.822}$                & $0.06$                                                  & $2.875$                                                     & $9.607$                                                     \\
			&  BAMM                    & $0.821$                & $0.058$                                                 & $\mathbf{2.868}$                                                     & $9.612$                                                     \\
			&  BAD                     & $0.806$                & $\mathbf{0.036}$                                                 & $2.917$                                                     & $\mathbf{9.615}$                                                     \\ \hline
			\rule{0pt}{2.5ex} \multirow{3}{*}{Suffix} &  Momask                  & $\mathbf{0.819}$                & $0.052$                                                 & $\mathbf{2.881}$                                                     & $9.659$                                                     \\
			&  BAMM                    & $0.814$                & $0.050$                                                 & $2.891$                                                     & $\mathbf{9.721}$                                                     \\
			&  BAD                     & $0.808$                & $\mathbf{0.044}$                                                 & $2.909$                                                     & $9.593$                                                     \\ \Xhline{1.5pt}
			&                                               &                        &                                                         &                                                             &                                                            
		\end{tabular}
	}
\vspace*{-0.15in}  \caption{\footnotesize Quantitative evaluation on temporal editing tasks on HumanML3D.}
\label{table:temporal-editing}
\vspace*{0.1in}
\end{table}

\textbf{\textit{Quantitative Comparison:}} Following~\cite{guo2022generating}, we report the metrics as the average over $20$ generation experiments, with a $95$\% confidence interval. We use $I=10$ iterations during the generation process. To demonstrate the core effectiveness of the proposed approach, we deliberately avoid  employing advanced VQ-VAE designs such as RVQ in the motion tokenizer.  Table~\ref{tab:no-rvq} shows that BAD, with a similar model size and design choices, consistently outperforms the  baselines, T2M-GPT~\cite{zhang2023generating}, an autoregressive motion model, and MMM~\cite{pinyoanuntapong2024mmm}, a mask-based generative motion model. By achieving the lowest FID score compared to T2M-GPT and MMM on both datasets, BAD demonstrates its ability to capture the sequential flow of information while simultaneously modeling rich bidirectional dependencies in complex motion sequences, indicating that the generated motions are natural and realistic. For text-motion consistency, BAD further improves R-Precision and MM-Dist metrics. In terms of inference speed, similar to MMM, BAD offers high inference speed compared to autoregressive \cite{zhang2023generating, zhong2023attt2m, jiang2024motiongpt} and diffusion-based motion models \cite{tevet2022humanmotiondiffusionmodel, zhang2022motiondiffuse, chen2023executing}.

Table~\ref{tab:rvq} compares BAD with two leading methods, Momask and BAMM, both of which use RVQ in their motion tokenizers, greatly improving the motion tokenizer metrics and consequently the overall framework. On HumanML3D, which is a larger and, therefore, more reliable dataset than KIT-ML, we achieve a better FID score compared to BAMM while remaining quite close to Momask. For text-motion consistency, our approach achieves comparable performance (R-Precision and MM-Dist) to both BAMM and Momask. Given that our pre-training approach can be easily adapted to other models, we anticipate that using an RVQ-based motion tokenizer could further improve our results, which we leave to future work.

\begin{figure}[t!]
	\centering
	\includegraphics[width=0.49\textwidth]{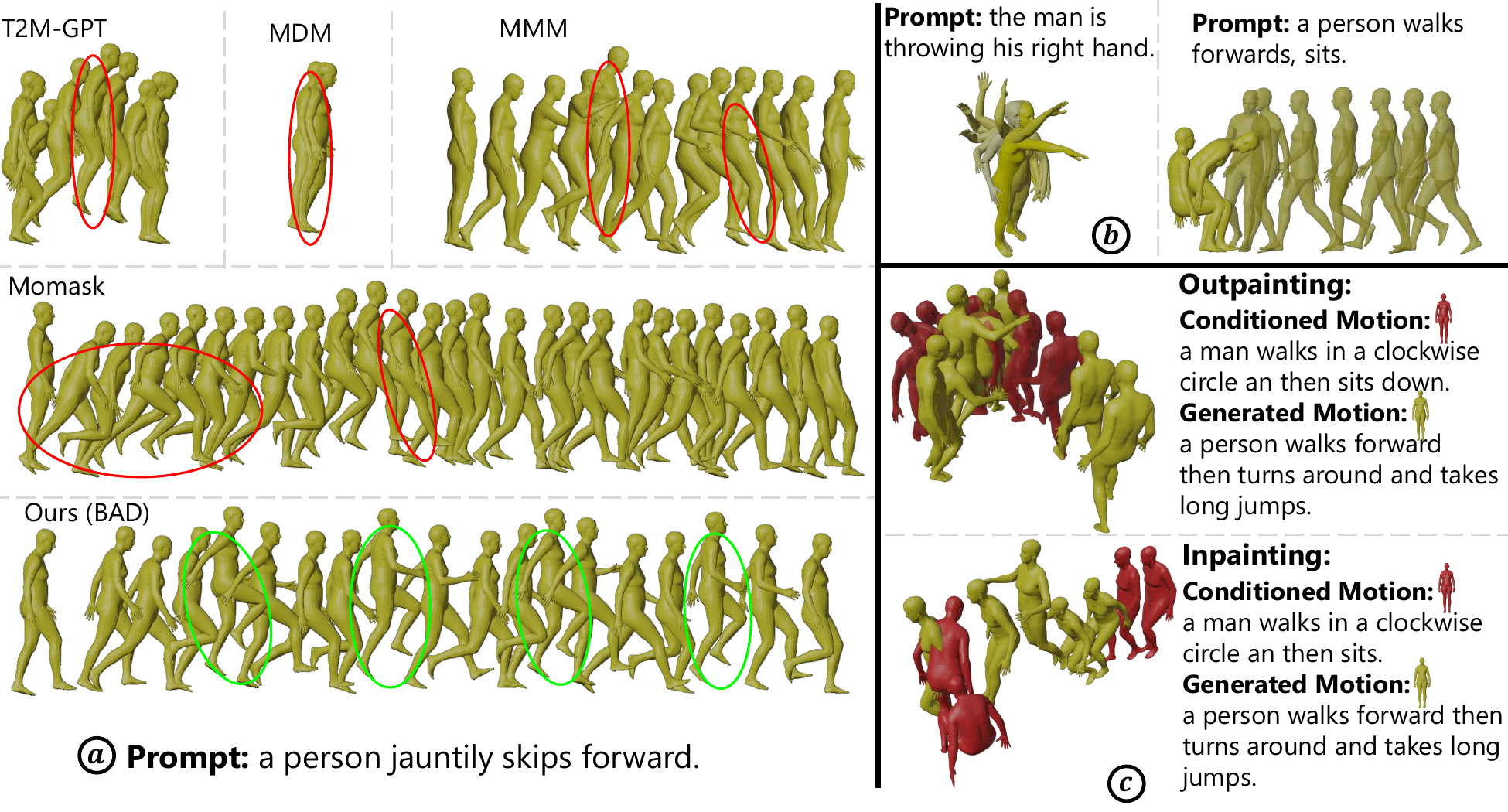}
	\caption{\footnotesize Quality Comparison. (a) Visualization of generated motions from various models for the same prompt, with \textcolor{red}{red} circles indicating defects and \textcolor{green}{green} circles highlighting correct, natural motions. (b) Additional motions generated by BAD. (c) Visualization of temporal editing tasks.}
	\label{fig:quality-comp-walk}
\end{figure}

We tested four temporal editing tasks on HumanML3D dataset: motion inpainting (generating the central $50$\% of a sequence conditioned on the first and last $25$\%), outpainting (generating the middle portion from the start and end of the sequence), prefix prediction (generating the second half of the sequence from the initial $50$\%), and suffix completion (generating the beginning of the sequence from the final $50$\%). These tasks are crucial for assessing motion sequence coherence and are illustrated in Fig.~\ref{fig:quality-comp-walk}(c), and Table~\ref{table:temporal-editing}. Results show that BAD outperforms advanced models Momask and BAMM in terms of FID score.

\vspace{.05in}
\noindent
\textbf{\textit{Qualitative Comparison:}} Fig.~\ref{fig:quality-comp-walk}(a) shows motions generated by different models for the same prompt. T2M-GPT and MDM fail to generate coherent motion, while MMM produces unnatural hand and foot movements, as indicated by the red circles. Momask initially generates a running motion, which is inconsistent with the prompt, and like MMM, fails to achieve natural hand and foot alignment. In contrast, BAD generates the motion with natural hand and foot movements and correctly performs the action multiple times.

\vspace*{-.15in}
\section{Conclusion}
\vspace*{-.1in}
We introduce BAD, a novel generative framework for text-to-motion generation, implemented in a two-stage process. First, a simple VQ-VAE is used to transform a raw 3D motion sequence into a sequence of discrete tokens. Next, a permutation-based corruption process corrupts the sequence, and a multi-layer transformer is trained to reconstruct it. By using a hybrid attention mask, our transformer captures rich bidirectional relationships while also learning causal dependencies between masked tokens. Extensive experiments demonstrate that BAD not only surpasses baseline approaches but also achieves competitive or superior results compared to RVQ-VAE-based models on various text-to-motion generation tasks. Notably, BAD can be easily adapted to other models and modalities, such as text, audio, and images.

\begingroup
\footnotesize
\bibliographystyle{IEEEbib}
\bibliography{strings}
\endgroup

\end{document}